\pdfoutput=1

\documentclass[11pt]{article}

\usepackage[final]{acl}

\usepackage{times}
\usepackage{latexsym}
\usepackage{booktabs}
\usepackage{adjustbox} 
\usepackage{placeins}
\usepackage[most]{tcolorbox}
\usepackage{enumitem}
\setlist{nosep,leftmargin=3mm}
\usepackage{xcolor}
\usepackage{caption}
\usepackage{subcaption}
\usepackage{array}
\usepackage{multirow}
\usepackage[T1]{fontenc}
\usepackage{bm}
\usepackage[utf8]{inputenc}

\usepackage{microtype}

\usepackage{inconsolata}

\usepackage{graphicx}

\title{Small LLMs Do Not Learn a Generalizable Theory of Mind via Reinforcement Learning}

\author{Sneheel Sarangi \\
  NYU Abu Dhabi \\
  \texttt{sneheelsarangi@nyu.edu} \\\And
  Hanan Salam \\
  NYU Abu Dhabi \\
  \texttt{hanan.salam@nyu.edu} \\}

\begin{document}
\maketitle

\begin{abstract}
Recent advancements in large language models (LLMs) have demonstrated emergent capabilities in complex reasoning, largely spurred by rule-based Reinforcement Learning (RL) techniques applied during the post-training. This has raised the question of whether similar methods can instill more nuanced, human-like social intelligence, such as a Theory of Mind (ToM), in LLMs. This paper investigates whether small-scale LLMs can acquire a robust and generalizable ToM capability through RL with verifiable rewards (RLVR). We conduct a systematic evaluation by training models on various combinations of prominent ToM datasets (HiToM, ExploreToM, FANToM) and testing for generalization on held-out datasets (e.g., OpenToM). Our findings indicate that small LLMs struggle to develop a generic ToM capability. While performance on in-distribution tasks improves, this capability fails to transfer to unseen ToM tasks with different characteristics. Furthermore, we demonstrate that prolonged RL training leads to models ``hacking'' the statistical patterns of the training datasets, resulting in significant performance gains on in-domain data but no change, or degradation of performance on out-of-distribution tasks. This suggests the learned behavior is a form of narrow overfitting rather than the acquisition of a true, abstract ToM capability.

\end{abstract}

\section{Introduction}
The ability to attribute mental states such as beliefs, desires, intentions to oneself and others, a capacity known as Theory of Mind (ToM), is a cornerstone of human social intelligence \cite{premack1978chimpanzee}. The development of artificial agents with a genuine ToM capability would represent a monumental leap towards more collaborative, predictable, and safe AI. The recent and rapid scaling of Large Language Models (LLMs) has ignited interest in their potential to develop such sophisticated social reasoning skills, with some models showing nascent ToM-like abilities on specialized benchmarks \cite{kosinski2023theory}. However, the question of whether LLMs possess a general-purpose human-like ToM capability remains contentious, 
\cite{shapira2023cleverhansneuraltheory}. Smaller language models, especially, struggle to perform well on existing benchmarks, lagging even when employed with mechanisms to boost performance on ToM tasks \cite{sarangi-etal-2025-decompose}.

Recently, there has been a paradigm shift in LLM training, where Reinforcement Learning (RL) has become a critical tool for unlocking capabilities beyond next-token prediction. Landmark models like DeepSeek-R1 have shown that RL with verifiable rewards (RLVR) can ``incentivize'' complex logical and mathematical reasoning, leading to skills that generalize to novel problems \cite{deepseekai2025deepseekr1incentivizingreasoningcapability}. Subsequent work, such as Logic-RL \cite{xie2025logicrlunleashingllmreasoning}, further demonstrated that targeted RL training on synthetic, rule-based tasks could foster a more abstract reasoning ability, transferable to different domains. Although recent work suggests that RLVR, when applied to ToM, can effectively boost ToM performance in LLMs \cite{lu2025theorymindbenchmarksneed}, the robustness and generalization of the gained capability remain unclear.
This raises a compelling question: \textit{Can the RL-driven success in the domain of logical reasoning be replicated for social reasoning? Specifically, can we use RL to train a small LLM to learn a generalizable ToM?}

In this work, we investigate this question by applying RL with verifiable rewards (RLVR) to small-scale LLMs, training them on a curated selection of ToM benchmark datasets. Previous studies have shown that LLM ToM capabilities may be attributed to learning shortcuts, heuristics, or spurious correlations \cite{shapira2023cleverhansneuraltheory} rather than a more general ToM capability. Similarly, we hypothesize that ToM capabilities learned by small models via RL may be brittle and fail to generalize. We suspect the models will learn to exploit dataset-specific statistical cues, thus ``hacking'' the performance metrics, rather than internalizing a coherent, abstract model of mental states.

To test this hypothesis, we train a small-scale LLM on various combinations of three prominent ToM datasets (HiToM \cite{wu-etal-2023-hi}, ExploreToM \cite{sclar2024exploretheorymindprogramguided}, FANToM \cite{kim-etal-2023-fantom}) and evaluate its zero-shot performance on a suite of held-out ToM tasks. Our contributions are threefold:
\begin{enumerate}
    \item We conduct a systematic empirical study on applying Reinforcement Learning with Verifiable Rewards (RLVR) to instill ToM in a small LLM, rigorously evaluating the generalization gap between in-distribution mastery and out-of-distribution performance.
    \item We provide direct evidence of statistical ``hacking,'' where prolonged RL training leads to inverted difficulty curves and negative transfer on varied-order ToM reasoning tasks, demonstrating that the model learns dataset artifacts rather than abstract principles.
    \item We demonstrate the extreme brittleness of the learned skills, showing that performance fails to transfer even to new task formats within the same data distribution, which underscores the superficial nature of the acquired capability.
\end{enumerate}

\section{Related Work}

\textbf{Machine Theory of Mind.} The development of computational systems exhibiting Theory of Mind (ToM), i.e. the capacity to attribute and reason about mental states, has been a persistent objective in artificial intelligence research \cite{pmlr-v80-rabinowitz18a}. Contemporary LLMs have exhibited substantial improvements, with performance metrics on established ToM benchmarks like ToMi \cite{le-etal-2019-revisiting} and BigToM \cite{gandhi2023understanding} approaching or exceeding human accuracy. Notwithstanding these advancements, the robustness of LLM-based ToM remains a subject of scrutiny, with previous studies pointing out that \cite{ullman2023large, shapira2023cleverhansneuraltheory} strong performance on ToM benchmarks may be an indicator that LLMs are using shortcuts or heuristics to answer questions. These concerns, alongside the saturation of existing benchmarks, have necessitated advancements in ToM evaluation methodologies. These include evaluations of higher-order ToM reasoning (e.g., iterated mental state attributions) \cite{wu-etal-2023-hi}, performance in naturalistic dialogue contexts \cite{kim-etal-2023-fantom}, and the creation of comprehensive datasets for evaluating a wider spectrum of ToM-related abilities. Recent benchmarks, such as OpenToM \cite{xu-etal-2024-opentom}, aim for more holistic assessments, for example, by evaluating LLMs' capabilities to understand mental states such as emotion, while others, such as ExploreToM \cite{sclar2024exploretheorymindprogramguided}, adversarially generate data to get a better measure of LLMs' ToM capabilities.

\textbf{Augmenting ToM in LLMs.} Recent research has proposed several distinct methodologies for enhancing the ToM capabilities of LLMs, primarily by introducing structured reasoning frameworks. SymbolicToM \cite{sclar-etal-2023-minding} employs LLMs to generate a symbolic graph representation of characters' belief states before addressing ToM queries. SimToM \cite{wilf-etal-2024-think}, inspired by Simulation Theory \cite{Shanton2010}, implements a two-stage process involving explicit perspective-taking by the LLM.  Similarly, Decompose-ToM \cite{sarangi-etal-2025-decompose} demonstrates that decomposing a complex ToM problem into a series of simpler, ToM-relevant sub-tasks can yield performance gains. However, these methods rely on external algorithmic control or predefined procedural frameworks to structure the LLM's inference process and remain dependent on the strength of the base model. For smaller base models, these methods do not significantly improve performance \cite{sarangi-etal-2025-decompose}. Additionally, ToM-related post-training methods can likely achieve a stronger upper bound in performance by directly injecting ToM capabilities into models. 

\textbf{Reinforcement Learning for LLMs}
The application of RL has fundamentally altered the trajectory of LLM development. Moving beyond the initial pre-training and supervised fine-tuning stages, RL allows models to be optimized directly for desired outcomes, such as helpfulness, harmlessness, or correctness \cite{ouyang2022traininglanguagemodelsfollow}. A pivotal innovation in this area is Reinforcement Learning from Verifiable Rewards (RLVR) \cite{lambert2025tulu3pushingfrontiers, deepseekai2025deepseekr1incentivizingreasoningcapability}. This technique sidesteps the ambiguity and cost of human feedback by using rewards derived from programmatic, rule-based, or otherwise verifiable outcomes.
This approach's success has been demonstrated by DeepSeek-R1, which showed that a pure RL training phase could dramatically boost performance on complex reasoning tasks in math and coding \cite{deepseekai2025deepseekr1incentivizingreasoningcapability}. The key insight was that by rewarding correct final answers, the model could be ``incentivized'' to develop its internal reasoning processes, which then generalized surprisingly well. Other works, such as Logic-RL \cite{xie2025logicrlunleashingllmreasoning}, which trained models on a corpus of synthetic logic puzzles, have shown that mastering these narrow, verifiable tasks led to improved performance on broader mathematical reasoning benchmarks, suggesting that the underlying logical principles were learned and transferred.
These successes in the domain of formal reasoning provide the direct motivation for our work. They establish a powerful precedent: RL can be used to cultivate abstract capabilities from specific, verifiable training data. This raises the question: ``To what extent can post-training techniques such as RL instill cognitive abilities like ToM in LLMs?'' While recent work has demonstrated positive results \cite{lu2025theorymindbenchmarksneed}, a comprehensive analysis of the nature and generalizability of potential ToM capabilities gained by these methods remains to be conducted. Thus, our work applies the successful RLVR methodology to the domain of ToM, investigating whether the same principles of emergent generalization hold for Theory of Mind.

\section{Methodology}

To investigate whether Reinforcement Learning with Verifiable Rewards (RLVR) can instill a generalizable Theory of Mind (ToM) in small-scale language models, we design a series of experiments that test both in-distribution performance and out-of-distribution generalization. Specifically, we use a 7B parameter model trained under different curriculum settings across curated ToM datasets.

We compile a suite of 4 ToM benchmarks encompassing a total of 12 tasks. These benchmarks were selected to span a wide range of input distributions, task formats, and levels of reasoning complexity. To evaluate generalization, we hold out one full benchmark (OpenToM \cite{xu-etal-2024-opentom}) and selected tasks from two others (FANToM \cite{kim-etal-2023-fantom} and HiToM \cite{wu-etal-2023-hi}) as evaluation-only datasets. This allows us to assess whether models trained on specific ToM data can transfer learned social reasoning capabilities to novel formats and tasks.

From the remaining datasets, we construct 7 training configurations by combining different subsets of the benchmarks. Each configuration serves as a distinct training regimen, enabling us to examine how the composition of training data affects learning and generalization. All models are trained using the RLVR framework, which optimizes for verifiable reward signals aimed at reinforcing logical reasoning behavior.

We then evaluate each trained model across all 12 ToM tasks, including both training-distribution and held-out tasks, to probe for signs of abstract and transferable ToM capabilities. The following subsections describe the datasets, training protocols, and RLVR implementation in more detail.

\subsection{Datasets}

\subsubsection{Training Datasets} 
We use three primary datasets for training: FANToM \cite{kim-etal-2023-fantom}, HiToM \cite{wu-etal-2023-hi}, and ExploreToM \cite{sclar2024exploretheorymindprogramguided}. These datasets were selected to capture a broad diversity of input formats, narrative styles, and Theory of Mind (ToM) challenges.
Specifically, FANToM comprises naturalistic dialogue conversations, HiToM features procedurally generated structured stories, whereas ExploreToM includes both narrative and adversarially structured false-belief tasks. For each dataset, we use 900 training samples, 300 validation samples, and 300 test samples.

\paragraph{Hi-ToM \cite{wu-etal-2023-hi}.}

HiToM evaluates higher-order ToM reasoning, extending up to fourth-order belief tracking. Inspired by the Sally-Anne paradigm \cite{BaronCohen1995}, it presents synthetic stories where characters enter, exit, and move objects between rooms. All stories are generated using templates, resulting in highly structured and consistent data. The core task is multiple-choice question answering with 15 answer options per instance. To assess generalization to higher-order reasoning, we exclude fourth-order questions from training and validation sets. Additionally, 10\% of examples are factual (no ToM required) to encourage grounding and reduce spurious policy learning.

\paragraph{FANToM \cite{kim-etal-2023-fantom}.}

FANToM presents ToM reasoning in naturalistic dialogue settings. Conversations feature characters joining and leaving dynamically, making belief tracking dependent on partial observability and turn-taking. From its suite of tasks, we use the binary false-belief classification task for training. To mitigate reward hacking and reinforce grounded reasoning, we augment the training set with true-belief and factual questions.

\paragraph{ExploreToM \cite{sclar2024exploretheorymindprogramguided}.}

ExploreToM is designed to challenge models with adversarially generated false-belief scenarios. It includes both structured (template-based) and narrative (LLM-infused) stories, focusing on nuanced belief modeling. From these, we use only the narrative stories to ensure diversity of input data. To ensure balanced learning, we sample the training data to include 70\% tasks requiring genuine ToM reasoning and 30\% solvable through simpler mental state tracking. This mix encourages the model to learn ToM capabilities beyond shallow pattern recognition.

\subsection{Evaluation Datasets}

To assess generalization, we evaluate model performance on three held-out datasets: (1) OpenToM \cite{xu-etal-2024-opentom}, (2) the FANToM List-response tasks \cite{xu-etal-2024-opentom}, and (3) the fourth-order HiToM task \cite{wu-etal-2023-hi}. These datasets are chosen to probe distinct generalization axes: narrative distribution shift, reasoning order extrapolation, and task format novelty. All three were excluded from training and validation to ensure a robust test of transferable ToM capability.


OpenToM consists of LLM-generated narratives inspired by the Sally-Anne false belief paradigm \cite{BaronCohen1995}, designed to evaluate both first- and second-order ToM reasoning. The dataset includes six core task types: coarse-grained location, fine-grained location, multihop-fullness, multihop-accessibility (each in first- and second-order forms), and an attitude task. Multihop tasks require two-step inference over belief chains, adding reasoning complexity beyond simple belief attribution. 

We use the extended version of OpenToM containing longer narratives, which better challenge narrative understanding and reasoning persistence. For evaluation, we sample 100 examples for each of the following tasks: first- and second-order variants of fine-grained location, multihop-fullness, and multihop-accessibility. To avoid label imbalance effects, we ensure an equal distribution of correct answer labels across samples.


We include two list-format tasks from the FANToM benchmark: *answerability-list* and *knowledge-awareness-list*. These tasks require the model to return a list of characters that meet a specified epistemic condition (e.g., knowing a fact, being able to answer a question), thereby testing multi-step reasoning under partial observability. Unlike the binary classification format used during training, these list-generation tasks evaluate the model's ability to generalize ToM reasoning to a different output structure and more complex aggregation logic.

\paragraph{HiToM (Fourth-Order) \cite{wu-etal-2023-hi}.}
To test generalization to higher-order ToM, we evaluate models on the fourth-order subset of HiToM. These examples require recursive reasoning about nested beliefs (e.g., ``A believes that B believes that C believes that D thinks...''), which were explicitly excluded from training. Performance on this task serves as a proxy for compositional ToM extrapolation.

\subsection{Reward Function Design}

To ensure consistency in model outputs and enable automated evaluation, we adopt a rule-based reward scheme inspired by prior work on logic-guided reinforcement learning \cite{xie2025logicrlunleashingllmreasoning}. The reward function is decomposed into two components: a \textit{format reward} and a \textit{correctness reward}, applied sequentially.

\paragraph{Format Reward.}
 We enforce a structured output format by requiring the model to enclose its intermediate reasoning within \texttt{<think>} and \texttt{</think>} tags, and its final answer within \texttt{<answer>} and \texttt{</answer>} tags. This constraint facilitates both reward parsing and model interpretability. The format reward \( S_{\text{format}} \) is defined as:

\[
S_{\text{format}} =
\begin{cases}
    0.1, & \text{if the output adheres to the} \\
         & \text{required format}\\
    0, & \text{otherwise}
\end{cases}
\]

\paragraph{Correctness Reward.}
If the format constraint is satisfied, we compute a correctness reward based on whether the model's extracted answer matches the ground truth. The correctness reward \( S_{\text{correct}} \) is defined as:

\[
S_{\text{correct}} =
\begin{cases}
    1, & \text{if the answer is correct} \\
    0, & \text{otherwise}
\end{cases}
\]

\noindent
The total reward for a response is the sum of the format and correctness rewards. This simple yet effective reward design allows us to decouple surface-level formatting from content correctness and encourages both structured reasoning and accurate answers.


\subsection{Training Algorithm: REINFORCE++}

We employ the REINFORCE++ algorithm \cite{hu2025reinforceefficientrlhfalgorithm} to optimize the language model using our rule-based reward signal. REINFORCE++ is a variant of the standard REINFORCE algorithm that omits the critic model used in Proximal Policy Optimization (PPO), thereby simplifying the training pipeline and reducing computational overhead.

Instead of using a learned value baseline, REINFORCE++ normalizes the reward across each training batch and uses this as a baseline to reduce variance in the policy gradient estimate. This approach has been shown to maintain strong sample efficiency and stable convergence without the additional complexity introduced by actor-critic methods in previous studies \cite{xie2025logicrlunleashingllmreasoning}.

\section{Experiments and Results}

\subsection{Experimental Setup}

We choose Qwen2.5-7B-Instruct for its strong instruction-following capabilities and growing adoption in RL-based LLM research, while remaining computationally feasible for systematic generalization studies with small models. We train a model for each combination of training sets from HiToM, FANToM, and ExploreToM, for a total of 7 trained models. We select the checkpoints to evaluate by picking the best-performing checkpoint on the validation set after training the models for 10 epochs. We use a batch size of 8, set the number of rollouts to 8, use a learning rate of $5e^{-7}$, and a temperature parameter of 0.6. We then conduct evaluations on all the considered datasets and tasks.

To investigate how gained capabilities and performance vary with the order of ToM, we experimented further with the HiToM dataset. In addition to our original model trained on Orders 1, 2, and 3, we trained six new checkpoints. Four of these were trained on single orders (1, 2, 3, and 4, respectively), and two were trained on combined orders (1 \& 2, and 1, 2, 3, \& 4). Each new model was trained on a dataset of 900 samples, drawn in equal proportions from its constituent orders. 

\newcommand{\tablesize}{\small}

\begin{table*}[t!]
    \centering
    \tablesize
    \setlength{\tabcolsep}{4pt}
    \caption{
        Performance comparison across all models. The highest score in each column is \textbf{bolded}. For compactness, column headers are abbreviated as follows: \textit{O1-O4} refer to the data test samples corresponding to HiToM reasoning orders (1st to 4th order); \textit{loc-fo}, \textit{loc-so}, \textit{full}, and \textit{acc} refer to OpenToM sub-tasks (location 1st order and 2nd order, fullness, and accessibility respectively); \textit{Ans} and \textit{Info} represent the FANToM List sub-tasks: Answerability and Information Access. All reported values are accuracy percentages (\%).
         Model names indicate the combination of datasets used during training: Hi = HiToM, Fan = FANToM, Exp = ExploreToM. For instance, \textit{Hi-Fan-Exp} denotes a model trained on all three datasets, while \textit{Hi-Fan} indicates training only on HiToM and FANToM.
    }
    \label{tab:updated_full_results}
    \newcolumntype{C}[1]{>{\centering\arraybackslash}p{#1}}
\begin{tabular}{l|c|c|ccccc|ccccc|ccc}
        \toprule
        \textbf{Dataset} & \textbf{ExpToM} & \textbf{FANToM} & \multicolumn{5}{c|}{\textbf{HiToM}} & \multicolumn{5}{c|}{\textbf{OpenToM}} & \multicolumn{3}{c}{\textbf{FANToM List}} \\
       \cmidrule(lr){1-1} \cmidrule(lr){2-2} \cmidrule(lr){3-3}\cmidrule(lr){4-8} \cmidrule(lr){9-13} \cmidrule(lr){14-16}
        \textbf{Model}& \textbf{All} & \textbf{All} & \textbf{All} & \textbf{O1} & \textbf{O2} & \textbf{O3} & \textbf{O4} & \textbf{All} & \textbf{loc-fo} & \textbf{loc-so} & \textbf{full} & \textbf{acc} & \textbf{All} & \textbf{Ans} & \textbf{Info} \\
        \midrule
        \textbf{Baseline}    & 60.5          & 20.5          & 40.6          & 49.2          & 41.7          & 35.8          & 35.8          & 55.3          & 76.0          & 43.0          & 52.5          & 53.9          & 29.6          & 44.4          & 14.8 \\
        \textbf{CoT}         & 57.5          & 27.0          & 44.4          & 65.8          & 48.3          & 29.2          & 34.2          & 59.2          & 79.0          & 44.0          & 56.3          & \textbf{61.6} & 43.0          & 48.0          & 38.0 \\
        \textbf{Hi}          & 56.9          & 18.5          & \textbf{82.9} & \textbf{73.3} & \textbf{77.5} & 86.7          & \textbf{94.2} & 59.9          & 76.0          & 42.0          & \textbf{64.7} & 57.6          & 45.8          & 50.0          & 41.6 \\
        \textbf{Fan}         & 54.4          & 91.5          & 41.7          & 72.5          & 37.5          & 25.8          & 30.8          & 59.9          & \textbf{90.0} & 41.0          & 57.6          & 56.9          & 40.9          & 38.3          & 43.4 \\
        \textbf{Exp}         & \textbf{85.1} & 14.5          & 37.1          & 59.2          & 41.7          & 25.0          & 22.5          & 60.0          & 79.0          & 45.0          & 59.3          & 58.8          & 43.2          & \textbf{55.6} & 30.8 \\
        \textbf{Hi-Fan}      & 59.5          & \textbf{93.0} & 71.7          & 67.5          & 67.5          & 73.3          & 78.3          & \textbf{61.8} & 83.0          & \textbf{46.0} & 60.3          & 60.8          & 44.0          & 46.8          & 41.2 \\
        \textbf{Hi-Exp}      & 83.2          & 24.0          & 81.2          & 70.0          & 75.8          & \textbf{89.2} & 89.9          & 61.2          & 79.0          & 45.0          & 62.3          & 59.8          & \textbf{45.9} & 51.2          & 40.6 \\
        \textbf{Fan-Exp}     & 79.0          & 91.0          & 42.5          & 60.8          & 41.7          & 35.8          & 31.7          & 56.9          & 74.0          & 39.0          & 58.8          & 55.4          & 43.6          & 41.2          & 46.0 \\
        \textbf{Hi-Fan-Exp}  & 81.1          & 92.0          & 81.2          & 70.8          & 76.7          & 86.7          & 90.8          & 59.4          & 75.0          & 45.0          & 58.8          & 59.8          & 41.8          & 34.8          & \textbf{48.8} \\
        \bottomrule
    \end{tabular}
\end{table*}

\subsection{Results}

\begin{figure*}[t!]
    \centering
    \begin{subfigure}[b]{0.7\textwidth}
        \centering
        \includegraphics[width=\textwidth]{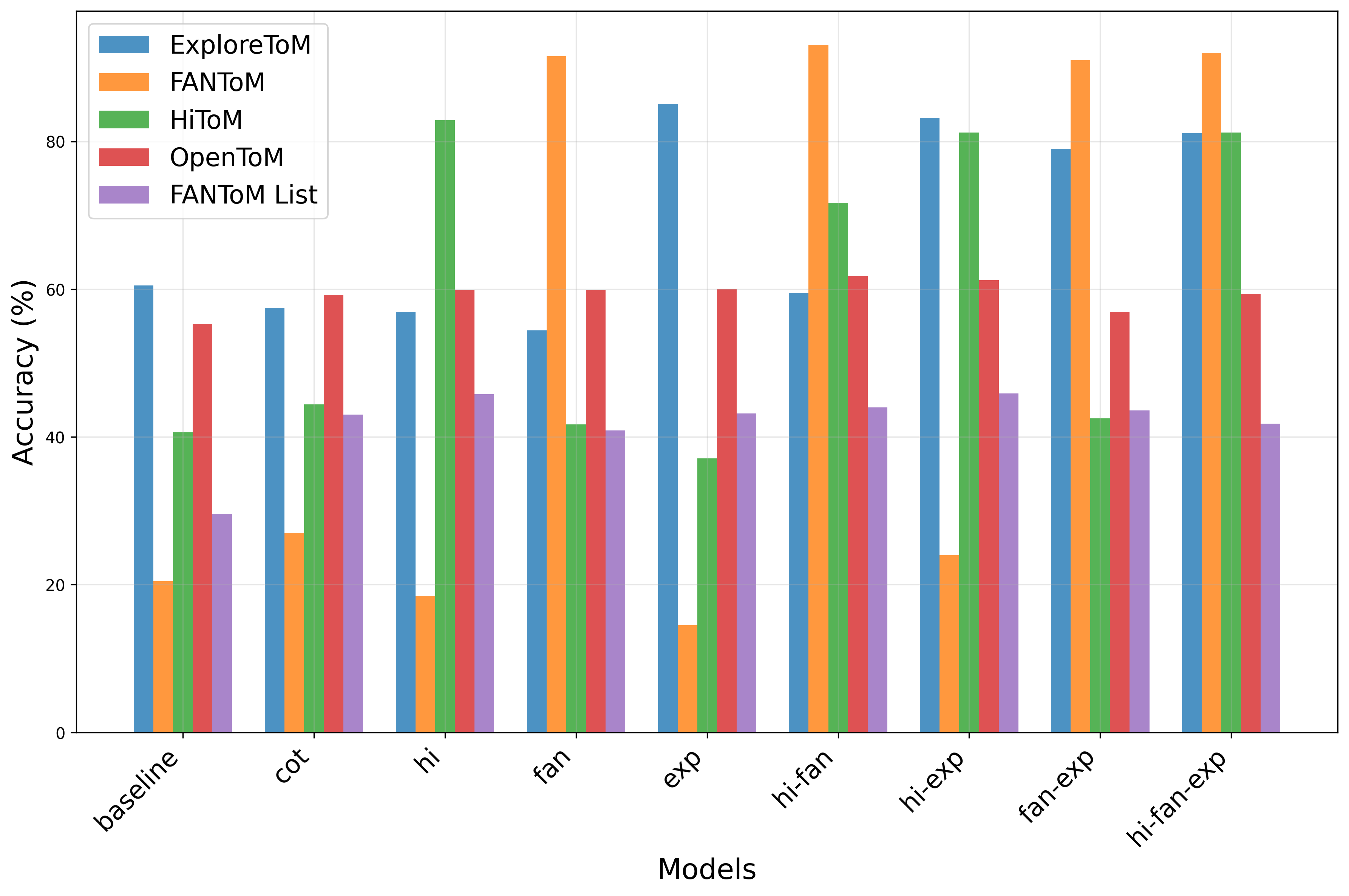}
        \caption{Overall Performance Comparison.}
        \label{fig:overall_perf}
    \end{subfigure}
    
    
    \begin{subfigure}[b]{0.48\textwidth}
        \centering
        \includegraphics[width=\textwidth]{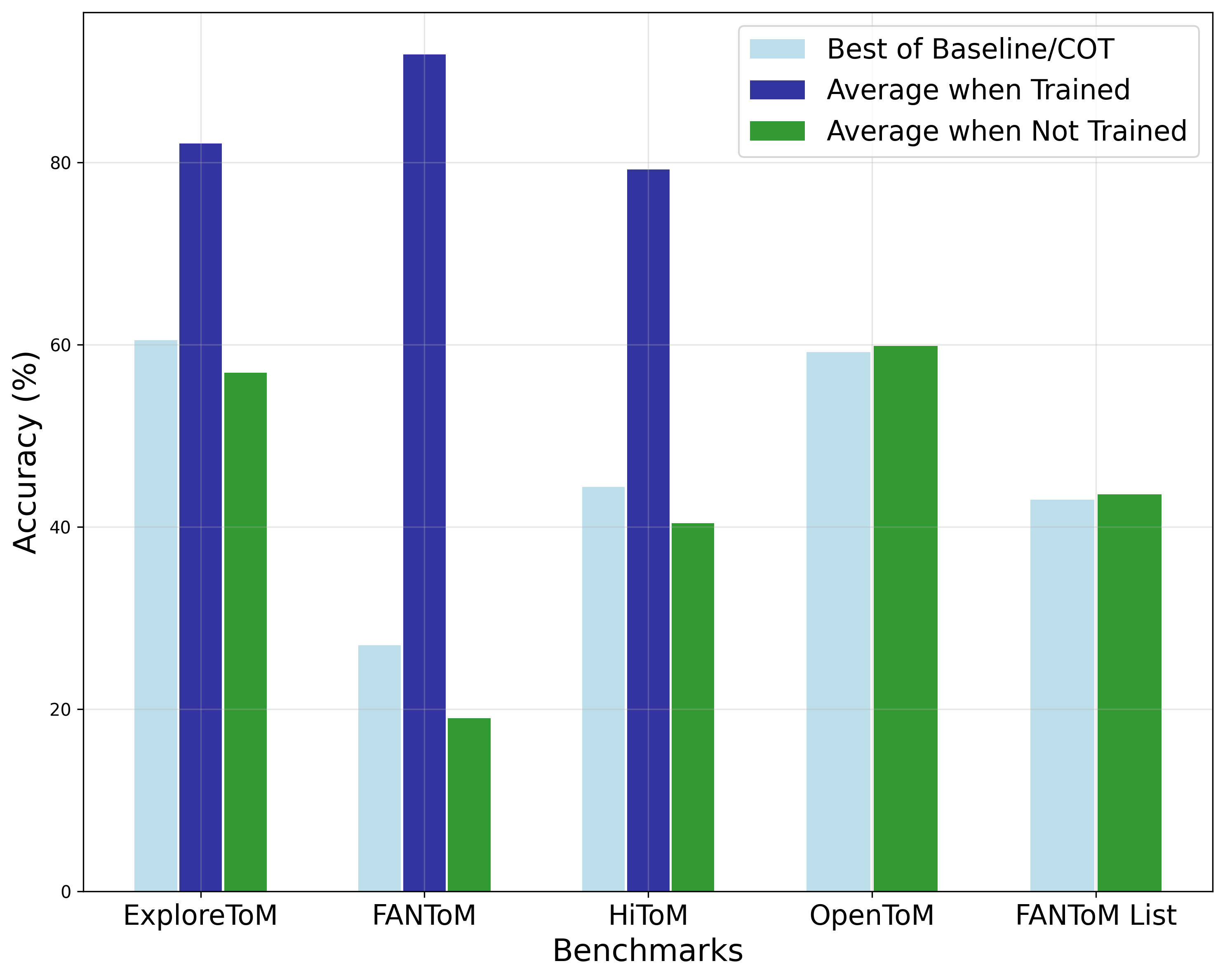}
        \caption{Baseline vs. Average Performance when trained on a dataset vs when not trained on the dataset.}
        \label{fig:baseline_vs_best}
    \end{subfigure}
    \hfill 
    \begin{subfigure}[b]{0.48\textwidth}
        \centering
        \includegraphics[width=\textwidth]{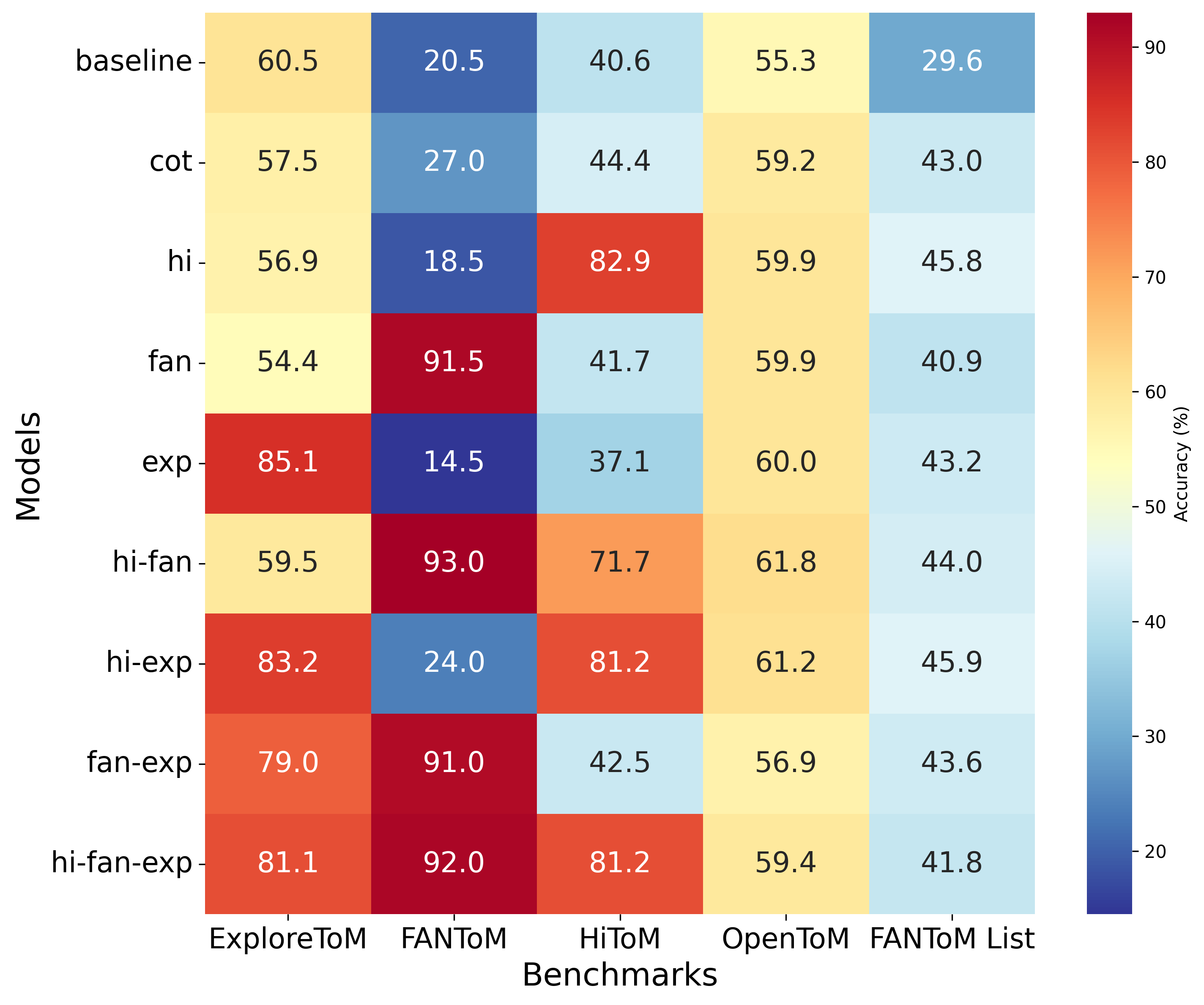}
        \caption{Performance Heatmap.}
        \label{fig:heatmap}
    \end{subfigure}
    
    \caption{
        \textbf{Summary of Model Performance.} 
        \textbf{(a)} A comparison of all models across benchmarks, showing high in-distribution scores.
        \textbf{(b)} A comparison highlighting the large performance gap between baselines and RL-specialized models on their target tasks. 
        \textbf{(c)} A heatmap visually representing the specialization of each model and its failure to generalize.
    }
    \label{fig:summary_results}
\end{figure*}

\subsubsection{RL Performance on In-Distribution Tasks}

RL training led to substantial performance improvements on in-distribution tasks, demonstrating its effectiveness for task-specific optimization. As demonstrated in \ref{fig:baseline_vs_best}, for all three of HiToM, FANToM, and ExploreToM, models trained on the datasets significantly outperform both baselines and models not trained on the datasets. Models trained on FANToM showed the largest improvements, outperforming the baselines by 65\%. HiToM trained models showed an improvement of 35\%, while ExploreToM trained models showed an improvement of 22\%. 

Additionally, this mastery extended to the specific reasoning styles of the training data. Models trained on the first to third-order reasoning tasks on the HiToM dataset also showed exceptional performance on the fourth-order reasoning tasks, gaining an accuracy increase of up to  59\%. Notably, this increase was greater than the performance improvement for the lower-order tasks, suggesting that the model learnt a policy that generalizes strongly to higher-order tasks. We analyze this phenomenon further in the Analysis.

\subsubsection{RL Performance on Out-Of-Distribution Tasks}

 Despite these impressive in-distribution gains, the models exhibited a critical failure to generalize to out-of-distribution (OOD) tasks. On the held-out \texttt{OpenToM} benchmark, the scores remained clustered in a tight range (56.9\% to 61.8\%) across all training regimens. No model significantly improved upon the chain-of-thought prompted performance of the untrained model with an accuracy of 59.2\%. For the FANToM List answering task, performance for the trained models similarly did not significantly improve past the chain-of-thought prompted base model's accuracy of 43\%, with the best performing model only obtaining an accuracy of 45.9\%. 

Overall, as shown in \ref{fig:baseline_vs_best}, the average accuracy of the trained models on the OpenToM and FANToM List tasks stayed close to the baseline performance. For the HiToM, FANToM, and ExploreToM tasks across training regimens not including the respective datasets, the performance was slightly lower than that of the base untrained model. In the worst cases, we observed a performance drop compared to the baselines, such as the accuracy on the FANToM task for the model trained on the ExploreToM dataset, which decreased to 14.5\% compared to the base model's accuracy of 27\%.

\begin{figure*}[h!]
    \centering
    \begin{subfigure}[b]{0.32\textwidth}
        \centering
        \includegraphics[width=\textwidth]{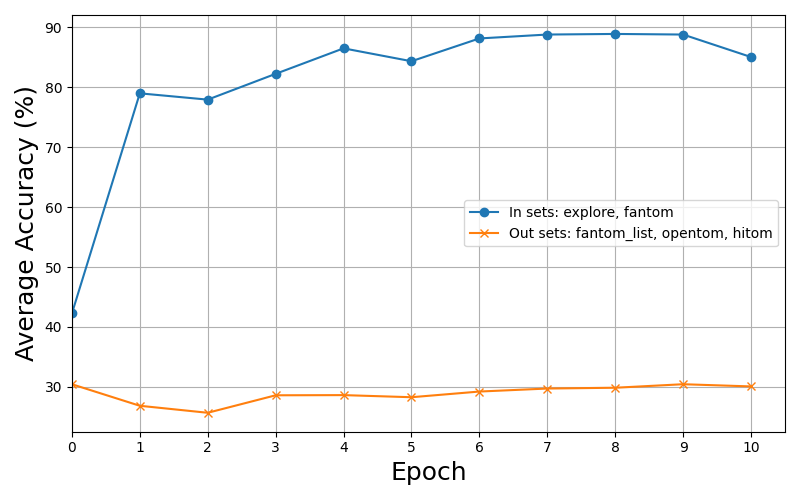}
        \caption{Trained on Explore, FANToM}
        \label{fig:epoch_exp_fan}
    \end{subfigure}
    \hfill 
    \begin{subfigure}[b]{0.32\textwidth}
        \centering
        \includegraphics[width=\textwidth]{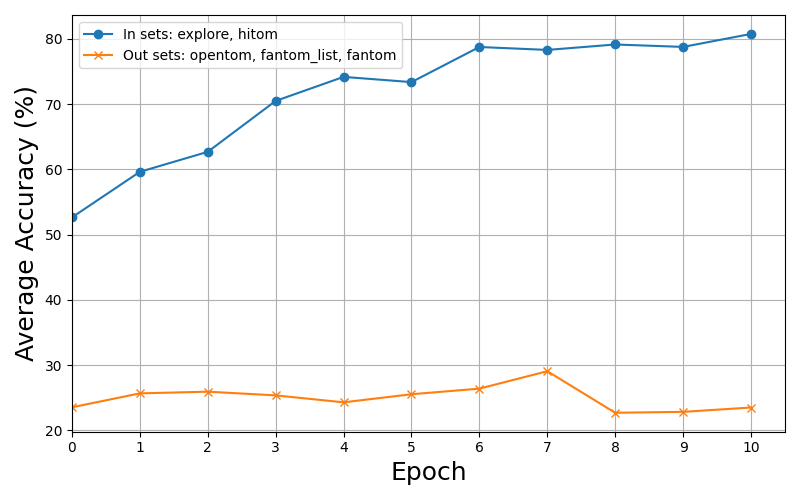}
        \caption{Trained on Explore, HiToM}
        \label{fig:epoch_exp_hi}
    \end{subfigure}
    \hfill 
    \begin{subfigure}[b]{0.32\textwidth}
        \centering
        \includegraphics[width=\textwidth]{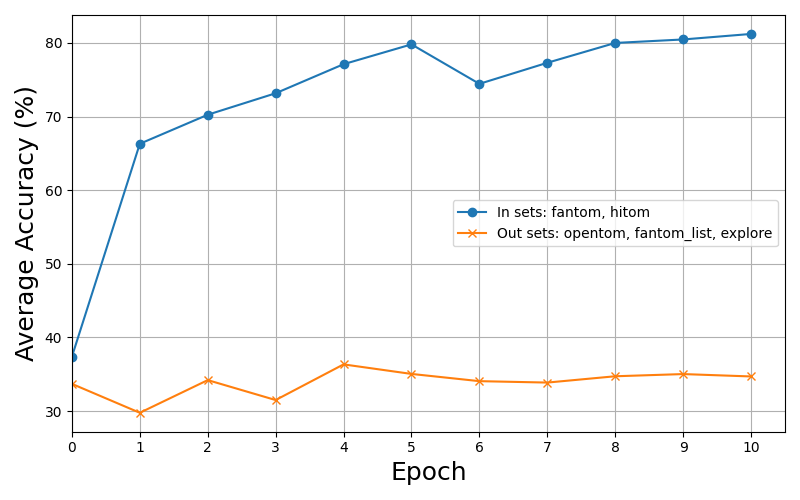}
        \caption{Trained on FANToM, HiToM}
        \label{fig:epoch_fan_hi}
    \end{subfigure}
    \caption{
        \textbf{Average Accuracy vs. Training Epoch.} These plots show a consistent divergence in performance between in-distribution sets (blue line, rising) and out-of-distribution sets (orange line, stagnating or falling).
    }
    \label{fig:acc_vs_epoch}
\end{figure*}
\begin{table}[h!]
\centering
\small
\caption{Performance accuracy (\%) of models trained on HiToM tasks of different orders on the overall HiToM benchmark. None is the baseline model, the following models are trained only on the orders mentioned.}
\label{tab:accuracy_orders}
\resizebox{\columnwidth}{!}{%
\begin{tabular}{@{} *{5}{lcccc} @{}}
\toprule
 & \multicolumn{4}{c@{}}{\textbf{Tested on}}\\
\cmidrule(l){2-5}
\textbf{Trained on} & \boldsymbol{$O_1$} & \boldsymbol{$O_2$} & \boldsymbol{$O_3$} & \boldsymbol{$O_4$} \\
\midrule
\textbf{None} & 65.8 & 48.3 & 29.2 & 34.2 \\
\boldsymbol{$O_1$} & 75.0 & 56.7 & 38.3 & 27.5 \\
\boldsymbol{$O_2$} & 41.7 & 67.5 & 76.7 & 70.8 \\
\boldsymbol{$O_3$} & 43.3 & 59.2 & 70.0 & 72.5 \\
\boldsymbol{$O_4$} & 35.0 & 52.5 & 73.3 & 85.8 \\
\boldsymbol{$O_{1,2}$} & 75.8 & 75.8 & 68.9 & 62.5 \\
\boldsymbol{$O_{1,2,3}$} & 73.3 & 77.5 & 86.7 & 94.2 \\
\boldsymbol{$O_{1,2,3,4}$} & 63.3 & 71.7 & 85.8 & 94.2 \\
\bottomrule
\end{tabular}
}
\end{table}

\subsubsection{Performance on Different ToM Orders}

To conduct a granular analysis of generalization within a single distribution, we evaluated models trained on specific reasoning orders from the HiToM dataset, with results detailed in Table \ref{tab:accuracy_orders}. The untuned baseline model exhibits a predictable difficulty curve, with accuracy degrading as cognitive load increases: it scores 65.8\% on first-order (O1) tasks, which falls to 48.3\% on O2, 29.2\% on O3, and 34.2\% on O4.

RL training, however, produces complex and non-intuitive patterns of generalization that reveal highly specialized, non-transferable strategies. Training on only O1 tasks, for instance, improves O1 performance to 75.0\% but fails to generalize upwards, causing a performance decrease on the O4 task to 27.5\%. Conversely, when trained exclusively on a single higher order (O2, O3, or O4), the model learns a strategy that is detrimental to the simplest case. This negative transfer is most severe when training only on O4, which drops O1 performance to 35.0\%, a nearly 31-point collapse from the baseline. Despite this, these specialized models perform well on their target and adjacent orders; the O3-trained model, for example, scores 70.0\% on O3 and 72.5\% on O4.

While single-order training reveals conflicting strategies, joint training on lower and higher order data in the training set can maintain performance while unlocking generalization. Training on O1 and O2 yielded a large improvement of over 30 percentage points on both O3 and O4 tasks compared to the baseline while maintaining performance on O1. This trend culminates in the model trained on orders 1, 2, and 3, which completely inverts the intuitive difficulty curve. It performs progressively better as the order increases (73.3\% on O1, 77.5\% on O2, 86.7\% on O3), achieving its peak accuracy of 94.2\% on the unseen fourth-order task. The inclusion of O4 data in the training set does not significantly alter these accuracies, indicating that performance had already saturated by exploiting patterns learned from the lower-order tasks.

\subsection{Performance On Task Variations}

We observe that models don't generalize to task variations even when the input data remains the same. The models trained on the false-belief task in the FANToM dataset do not outperform baselines on the list answering tasks. Training on only the FANToM dataset slightly reduced the performance on the list-answering task by 2.1\%, whereas the models trained jointly on the HiToM or ExploreToM datasets only outperformed the baselines by <1\%.

\subsubsection{Training Behavior Analysis}
\textbf{Accuracies on out-sets remain stagnant through training runs.} To better understand how model behavior changes through a training run, we plot in/out set accuracies over training epochs in Figure \ref{fig:acc_vs_epoch}. We observe that while the in-set accuracies consistently increase, out-set accuracies stay stagnant with no significant changes. This serves as further evidence that models overfit to perform better at in-distribution tasks.

\section{Discussion}
\textbf{In-Distribution Mastery Does Not Translate to Out-of-Distribution Generalization.} The primary finding of this work is the stark discrepancy between a model's ability to master a specific ToM benchmark and its ability to generalize that skill. Our experiments consistently show that RLVR is an exceptionally effective optimizer for in-distribution tasks, with performance on datasets like FANToM and HiToM increasing by over 40-60 percentage points post-training (Table \ref{tab:updated_full_results}). This confirms the power of RL in achieving high scores on a given benchmark. However, this success is purely local. When these specialized models were evaluated on the held-out OpenToM benchmark, their performance was indistinguishable from the untuned baseline. This suggests the learned "skill" is inextricably tied to the source distribution, preventing transfer and indicating the absence of an abstract, generalizable capability. This outcome provides strong empirical support for concerns raised by prior work \cite{shapira2023cleverhansneuraltheory, ullman2023large} that strong benchmark scores can be misleading.

\textbf{Training Dynamics Reveal a Divergence Toward Overfitting.} The analysis of model performance over training epochs provides a clear mechanism for this failure to generalize. As shown in Figure \ref{fig:acc_vs_epoch}, the learning curves for in-distribution and out-of-distribution datasets diverge. In-distribution accuracy steadily rises as the model is rewarded for correct answers, while out-of-distribution accuracy remains stagnant. This pattern is a classic signature of overfitting, where the model progressively learns the statistical idiosyncrasies and spurious correlations of its training data rather than the underlying principles of the task. This outcome contrasts sharply with findings in the logical reasoning domain \cite{deepseekai2025deepseekr1incentivizingreasoningcapability}, suggesting that the ambiguity and contextual nuance inherent to social reasoning tasks may make their benchmarks more susceptible to this kind of statistical exploitation via RL.

\textbf{Inverted Difficulty Curves Suggest Hacking of Dataset Artifacts.} The experiments on HiToM's tiered reasoning orders offer the most compelling evidence of "hacking" rather than learning. A model possessing a genuine ToM capability should find higher-order reasoning more difficult, a trend observed in our baseline model. Instead, the RL-trained model inverted this difficulty curve, performing best on the unseen and most complex fourth-order task (Table \ref{tab:accuracy_orders}). This paradoxical result is highly unlikely to stem from a sudden mastery of complex recursive thought. A more plausible explanation is that the model identified and exploited structural artifacts in the templated HiToM data that become more pronounced or predictive in higher-order examples. This finding serves as a cautionary tale about the face validity of benchmark performance, as the model's highest score was achieved through a method contrary to the intended reasoning path.

\textbf{Learned Skills are Brittle to Changes in Task Format.} Beyond failing to generalize to new datasets, the learned capabilities were also brittle to changes in task format within the same dataset. A model that achieved over 90\% accuracy on FANToM's binary false-belief questions showed no meaningful improvement on the FANToM List tasks, despite both tasks relying on the same conversational context and underlying mental state information. This demonstrates that the model did not learn a flexible internal representation of the characters' beliefs that could be queried in different ways. Instead, it learned a rigid policy for a specific (context, question type) $\rightarrow$ answer mapping. The inability to handle slight variations in the query format underscores the superficiality of the learned skill, which lacks the robustness expected of a true cognitive capability.

\section{Conclusion}
In this paper, we investigated whether Reinforcement Learning with Verifiable Rewards (RLVR), a technique successful in fostering logical reasoning, could be used to instill a generalizable Theory of Mind (ToM) in small-scale language models. By training a 7B parameter model on various combinations of prominent ToM benchmarks and evaluating on a suite of held-out tasks, we sought to determine if the model could acquire an abstract and transferable social reasoning capability.

Our findings demonstrate that while RLVR led to dramatic performance increases on in-distribution datasets, this specialized mastery failed to generalize. Across all training regimens, model performance on unseen ToM benchmarks and novel task formats remained stagnant, showing no significant improvement over a simple baseline. We presented further evidence that prolonged RL training encourages models to overfit to the statistical artifacts of the training data, a phenomenon we term as ``hacking''. This was most evident in the paradoxical finding that a model trained on lower-order reasoning tasks performed best on a more complex, unseen higher-order task, suggesting it had exploited structural patterns in the data rather than learning the underlying cognitive principle.

We conclude that, for small LLMs, the application of RLVR on current ToM benchmarks does not lead to the emergence of a genuine, general-purpose ToM. The learned behaviors are narrow, brittle, and indicative of sophisticated pattern matching rather than abstract social intelligence. These results underscore the limitations of current evaluation paradigms and suggest that developing truly socially intelligent AI will require advancements beyond optimizing for correct answers on existing benchmarks, potentially involving more robust and diverse training data or novel reward mechanisms that can assess the fidelity of the reasoning process itself.

\bibliography{custom}

\newpage

\end{document}